# Structural Analysis of Hindi Phonetics and A Method for Extraction of Phonetically Rich Sentences from a Very Large Hindi Text Corpus

Shrikant Malviya*, Rohit Mishra† and Uma Shanker Tiwary‡
Department of Information Technology
Indian Institute of Information Technology, Allahabad, India 211012
*Email: shrikant.iet6153@gmail.com
†Email: rohit129iiita@gmail.com
‡Email: ustiwary@gmail.com

*Abstract*—Automatic speech recognition (ASR) and Text to speech (TTS) are two prominent area of research in human computer interaction nowadays. A set of phonetically rich sentences is in a matter of importance in order to develop these two interactive modules of HCI. Essentially, the set of phonetically rich sentences has to cover all possible phone units distributed uniformly. Selecting such a set from a big corpus with maintaining phonetic characteristic based similarity is still a challenging problem. The major objective of this paper is to devise a criteria in order to select a set of sentences encompassing all phonetic aspects of a corpus with size as minimum as possible. First, this paper presents a statistical analysis of Hindi phonetics by observing the structural characteristics. Further a two stage algorithm is proposed to extract phonetically rich sentences with a high variety of triphones from the EMILLE Hindi corpus. The algorithm consists of a distance measuring criteria to select a sentence in order to improve the triphone distribution. Moreover, a special preprocessing method is proposed to score each triphone in terms of inverse probability in order to fasten the algorithm. The results show that the approach efficiently build uniformly distributed phonetically-rich corpus with optimum number of sentences.

## I. Introduction

When a set of sentences would be called as phonetically-rich set? The answer to the question depends on two statistical properties of phonetic distribution. First one is to know about the characteristic distribution of phonemes which decides the phonetic richnes of a sentence. On the other hand second property talks about the phonetic resemblance between extracted sentences and language in study. Evidently, this field of study is significantly related to automatic speech recognition (ASR) and speech synthesis(TTS) [1].

The task of extracting phonetically rich and balanced sentences involves, analyzing a large corpus and performing the procedure to extract sentences using sentence extraction criteria, based on various stochastic methodologies [2]. Additionally, a corpus has to be phonetically balanced through made up of sentences having phonetic units as per its distribution in the natural spoken language. Now based on these set of sentences which contains all the phonemes in most of the possible contexts, when recorded, would produce a phonetically rich and balanced speech corpus [3], [4].

Compare to some studies which are based on employing words, syllables and monophones, most of the current research in the development of Automatic Speech Recognition (ASR) and Text to Speech (TTS) systems widely focused on how the contextual phone units e.g. triphones and diphones could be used for improving the robustness of the systems [5], [6].

The field of constructing a phonetically-rich sentence corpus is relevant to various applications i.e. ASR and speech synthesis and many more. For instance, to estimate a robust acoustic model, a phonetically rich speech dataset is a basic requirement [7]. Phonologists have found useful to have such a specific corpora in order to develop a sytem to analyze speech production and variability [8]. In speech therapy, to find communicative disorders of patients, phonetically-rich sentences are often utilized in assessment of patient's speech production under surveillance in various phonetic/phonological contexts [9].

A novel approach based on the triphone distribution has been formulated and evaluated in this paper. The problem could be elaborated formally as: suppose a corpus $C$ is given which consists of $s$ sentences, find a subset $K$ such that subset $K$ containing $s_k$ sentences having uniformly distributed triphones. In first appearance the problem looks simple, but due to inherent high computational time complexity, the task is considered to be approached diligently. The problem is a non-polynomial type as it generates a set of instances based on the combinations of triphones and should be considered as an intractable problem [10].

A two phase algorithm has been proposed to extract set of phonetically-rich triphone sentences from EMILLE corpora in order to build a phonetically-rich corpus for Hindi. Hindi corpus from the EMILLE has been selected for the experiments and evaluation of the proposed method [11]. Furthermore, a greedy distance based criteria is employed in the algorithm for the selection of sentences through a finite number of iterations.

## II. Characteristic Structure of Hindi Language

### A. Phonetic Structure

After Mandarin, Spanish and English, Hindi is the most natively spoken language in the world, almost spoken by 260 million people according to Ethnologue, 2014 [12]. It is recognized as an official language of India in Devanagari script. Hindi is a direct descendant of Sanskrit through Pakrit (One of the Ancient Indo-Aryan language) and Apabhransha (Corruption or corrupted speech). Additionally, it is considered to be influenced by Dravidian language (Telugu, Tamil, Kannada and Malayalam spoken mostly in southern India), Turkish, Persian, Arabic, Portuguese and English.

In Hindi, there are total 33 consonants and 13 vowels used for speaking as well as writing. All the vowels have two forms of writing. First one is standalone form, being used as it is shown in the table I. For example in the Hindi word आम ('mango') contains vowel आ '/a:/' is being used in the same way as defined. The other one is mātrā form, which worked as a consonants modifier. For example in the word दान ('donation') [/d a: n/], vowel आ '/a:/' is attached with the consonant द /'d'/. Table I and III show a complete list of Hindi vowels and consonants with their phonetic details. For the convenience of representation PLU, phone like unit has been defined for each letter.

On the basis of place of origin, consonants are of categorized in the following four categories :

1) Regular Consonants (25)
2) Semivowels Consonants (4)
3) Sibilant Consonants (3)
4) Fricative Consonants (1)

Consonants are being articulated and amalgamated with vowels by stopping the air moving out of the mouth. Based on this fact of obstruction, the consonants are divided into five groups (वर्ग [vargas]). These vargas (groups) are ordered according to where the tongue is located in the mouth. As we move forward in the list of consonants groups, the tongue position is also move forward inside the mouth during the articulation. Table II clearly distinguishes this categorization of regular consonants. Each group (vargas) has five consonants ordered in a specific way according to how they are realized during the articulation. The members of each group are ordered in the sequence of voiceless unaspirated, voiceless aspirated, voiced unaspirated, voiced aspirated and nasal. Thus a sort of 5x5 matrix has been formed here to represent all the regular consonants table II.

Table I
A Glance On Hindi Vowels With Example and Phonetic Presentation [13]

| Hindi Orthography | IPA Symbol | PLU Symbol | Example |
|---|---|---|---|
| अ | /ə/ | A | अध्यापक 'Teacher' |
| आ | /a:/ | AA | आम 'Mango' |
| इ | /i/ | I | इत्र 'Perfume' |
| ई | /i:/ | II | ईश्वर 'God' |
| उ | /u/ | U | उत्तर 'Answer' |
| ऊ | /u:/ | UU | ऊष्मा 'Heat' |
| ए | /e:/ | E | एक 'One' |
| ऐ | /æ:/ | EE | ऐनक 'Specs' |
| ओ | /o:/ | O | ओस 'Dew' |
| औ | /ɔ:/ | OO | औषधि 'Medicine' |
| अं | /aŋ/ | MN | अंगूर 'Grape' |
| अः | /əh/ | AH | अतः 'So' |
| ऋ | /ɽ/ | RI | ऋषि 'Sage' |

Table II
Regular Consonants Categorized According to Tongue Position during Articulation

| Group Name | Tongue Position | Group Members | | | | |
|---|---|---|---|---|---|---|
| | | VL* | VLA# | V¶ | VA§ | N† |
| क वर्ग | Velar | क | ख | ग | घ | ङ |
| च वर्ग | Palatal | च | छ | ज | झ | ञ |
| ट वर्ग | Retroflex | ट | ठ | ड | ढ | ण |
| त वर्ग | Dental | त | थ | द | ध | न |
| प वर्ग | Labial | प | फ | ब | भ | म |

*VL = VOICE LESS
#VLA = VOICE LESS ASPIRATED
¶V = VOICED
§VA = VOICED ASPIRATED
†N = NASAL

Table III
A Glance On Hindi Consonants With Example and Phonetic Presentation [13]

| Hindi Orthography | IPA Symbol | PLU Symbol | Example |
|---|---|---|---|
| क | /k/ | K | कर 'Tax' |
| ख | /kʰ/ | KH | खेल 'Sport' |
| ग | /g/ | G | गाय 'Cow' |
| घ | /gʱ/ | GH | घर 'House' |
| ङ | /ŋ/ | WN | पङक 'Mud' |
| च | /tʃ/ | CH | चमत्कार 'Miracle' |
| छ | /tʃʰ/ | CHH | छाया 'Shadow' |
| ज | /dʒ/ | J | जल 'Water' |
| झ | /dʒʱ/ | JH | झरना 'Waterfall' |
| ञ | /ɲ/ | ZN | कञ्चे 'Marbles' |
| ट | /ʈ/ | TT | टहलना 'Amble' |
| ठ | /ʈʰ/ | TTH | ठीक 'Fine' |
| ड | /ɖ/ | DD | डाक 'Post' |
| ढ | /ɖʱ/ | DDH | ढोल 'Drum' |
| ण | /ɳ/ | AN | गणना 'Calculation' |
| त | /t/ | T | तरबूज 'Watermelon' |
| थ | /tʰ/ | TH | थाना 'Station' |
| द | /d/ | D | दर्पण 'Mirror' |
| ध | /dʱ/ | DH | धातु 'Metal' |
| न | /n/ | N | नमक 'Salt' |
| प | /p/ | P | परोपकार 'Charity' |
| फ | /pʰ/ | PH | फल 'Fruit' |
| ब | /b/ | B | बरसात 'Rain' |
| भ | /bʱ/ | BH | भक्ति 'Devotion' |
| म | /m/ | M | मदद 'Help' |
| य | /j/ | Y | यंत्र 'Instrument' |
| र | /ɾ/ | R | रक्षा 'Defence' |
| ल | /l/ | L | लगन 'Diligence' |
| व | /ʋ/ | V | वास्तु 'Architectural' |
| स | /s/ | S | समुदाय 'Community' |
| ष | /ʂ/ | SHH | धनुष 'Bow' |
| श | /ʃ/ | SH | शक्ति 'Power' |
| ह | /ɦ/ | H | हल 'solution' |

ʰVOICELESS ASPIRATED
ʱVOICED ASPIRATED

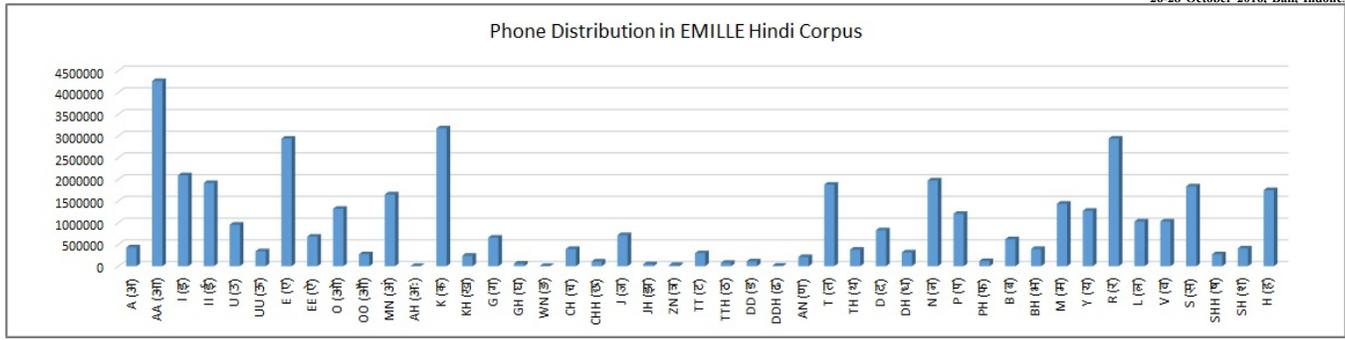

Figure 1. Distribution of Various Phones in EMILLE Corpus

In addition to the regular consonants, Hindi language consists of four semivowels consonants, three sibilant consonants and one fricative consonants. The four semivowels are य, र, ल and व. य, and ल are almost pronounced similarly as their counterpart 'Y' and 'L' pronounced in English. व has a flexible pronunciation as per the context, regional influence. It is sometimes articulated similar to 'W' like in स्वामी (Swamy). Other times it is pronounced closer to the english 'V'. Rest four consonants are of fricative type. Among them three श, ष, स are pronounced in voiceless manner. ह is fricative voiced but mostly pronounced similar to english 'H' spoken from the glottis.

*B. Statistical Analysis Hindi Phonetic Units*

In order to analyze and observe the statistical distribution of Hindi phonetic units, EMILLIE[1] (Enabling Minority Language Engineering) corpus is used here, which was developed at the Department of Linguistics, Lancaster University to encourage research on the South Asian Languages (Indic Languages, Chinese Language). The corpora consists of three types of sub-corpora: monolingual, parallel and annotated. Among them, it has fourteen monolingual corpora in both form written as well as spoken(for some languages) in all the fourteen South Asian languages generally spoken in countries India, Pakistan, Bangladesh, Nepal and Sri Lanka. Parallel corpus are also provided in the form of texts in English accompanied with its translation in Hindi, Punjabi, Gujarati and Urdu [11].

The Hindi corpus taken from EMILLE, is composed of the content retrieved from websites IndiaInfo, Ranchi Express Webdunia and some provided from CIIL[2]. Overall the corpus contains approximately a total of 12.39 million words. Other features of the corpus are shown in table IV. Further, figure 1 is presenting a detailed picture of how the various Hindi phones are weighted in terms of occurrence in the corpus. It is clearly visible that certain vowels and consonants are used in abundance in the written and spoken form of Hindi. Perhaps characters from क (/k/) to ण (/an/) have very less occurrence comparatively.

[1]http://catalog.elra.info/product_info.php?products_id=696
[2]http://www.ciil.org/

Table IV
STATISTICS OF EMMILLE HINDI CORPUS

| Feature of Data Set | Data |
|---|---|
| Total Sentences | 1,784,784 |
| Total Words | 12,390,000 |
| Unique Words | 754,648 |
| Min length (Word) | 4 |
| Max Length (Word) | 18 |

### III. RELATED WORK

In the last few years, significant efforts have been put forwarded in building and developing a speech corpora for many languages. These efforts have fetched out the enmeshed relationship which exists between the written and spoken form of a language. Although, the relation between the both mode of language representation is essential to be understood as both are the necessary part for several applications e.g. automatic speech recognition (ASR), text to speech (TTS), speech therapy and variability analysis.

In [2], phonetically-rich sentences are extracted from the large corpus of Hindi language. The approach considers syllables as a basic unit to compare phonetically-rich sentences and justify the fact that the phonetically-rich sentence set should have the same distribution of syllables as in the corpus. A syllable is described as a smallest segment of an utterance, which can be pronounced in isolation by a native speaker. For example consonant 'क' will form total 13 syllables by applying each vowel on it e.g. "क, का, कि, की, कु, कू, के, कै, को, कौ, कं, कः, कृ".

It has been found out by exploring the linguistic characterization of speech and text that text representation of a language is more explicit, more structurally complex and elaborate, more planned and organized than the counterpart speech [14]. Due to these contrast distinction it's been clearly envisaged that the written form of corpora has to be designed first before creating and recording spoken form of it. Therefore, the role of linguists and phoneticians is to build diligently phonetically-rich and phonetically-balanced written corpora before handing it to the specialist of speech recording for the recording purpose.

According to [4], even though each speaker has his/her own speaking style, but for the same language their speech has a common phonological structure. This fact encourages linguists of many languages to select phonetically rich and balanced text and speech corpora. In [15], creation of phonetically rich as well as balanced text corpora is starts with selecting sentences and phrases which are build of mostly phonetically rich words. These selected phrases and sentences further checked and verified for the balanced distribution of phonological units. Such text corpora then forwarded for the recording in order to construct the speech dataset needed to develop an efficient ASR (Automatic Speech Recognizer).

Phonetically balanced and rich property is on a prime focus for developing a speech corpora since a long time [16]. The reason is that most of these corpus are designed in order to train a speech recognizer. But, for other kind of research i.e. prosody based speech analysis, such databases would not be helpful. Hence, in [17] a specific speech corpus is designed with covering all the prosodic variation for Mandarin based on the assumption that prosodic properties are determined by the sentence types as the sentence type defined intonation in the speech.

## IV. Proposed Approach

### A. Triphones : extraction & analysis

Fundamentally, a triphone is a sequence of three phonemes. Inevitably, this is equivalent to say that it is a 3rd order Markov Chain or a trigram over phonemes. From the Markov assumption point of view, it could be said as the probability of the third phoneme given the first two phoneme. In other words, in a chain of long utterance this model will consider all previous phonemes irrelevant other than the preceding two.

Table V
A Comparison of Monophone And Triphone Transcription of a Hindi Word

| Word | Monophone Form | Triphone Form |
|---|---|---|
| चींटी | [tʃ i: ʈ i:] | [#-tʃ-i: tʃ-i:-ʈ i:-ʈ-i: ʈ-i:-#] |

Formally, a triphone is expressed in a sequence i.e. $(pl-p-pr)$. Here $pl$ characterizes the phone which comes before $p$ and $pr$ is one which follows it. Table V shows how monophones and triphones sequence of the word "चींटी" (Ant) would be represented.

By observing the capability of characterizing the surrounding environment of a given phoneme, contextual phone units i.e. triphones are voluminously applied to speech technology and made immense impact in the development of speech recognition and synthesis. Above explained reasons motivate us to select and explore the triphone as a fundamental unit of analysis for the proposed algorithm in order to build a phonetically rich corpus.

### B. Phonetic Richness Metric Used

Before, discussing the process of extracting phonetically rich sentences, a metric for measuring phonetic richness of a sentence needed to be formalized in the expedition of solving the problem. As a greedy algorithm is being used for the sentence selection based on the triphone they contained, a probabilistic metric is devised to pre-select and rank the sentences first, based on the triphone density it has. The metric takes the probability of each triphone exists in the corpus in order to select the high rank sentences by favoring the sentences consist of least frequent triphones over the sentences which consist of most frequent triphones. It's been implemented through performing summation of the reciprocal probability (calculated on the corpus) of each triphone belong to the sentence.

Formally, the complete scenario can be described more clearly in the subsequent way. View, the corpus $K$ comprised of a set of sentences $S = \{s_1, s_2, s_3, ..., s_n\}$. Consider each sentence $s$ is made of sequence of $m$ triphones in a particular order, represented as $T = \{t_1, t_2, t_3, ..., t_m\}$. As the richness of a sentence depends upon the distribution of triphones across it, so the triphones have to be scored first. Hence, a priori probability of each triphone is calculated as $P_K(t_i)$ : probability of $t_i^{th}$ triphone in corpus $K$. $P_K(t_i)$ is calculated by:

$$P_K(t_i) = \frac{Count(t_i)}{\sum_{j=1}^{j=m} Count(t_j)}$$

Where $Count(t_i)$ shows the total frequency of $t_i^{th}$ triphone in the corpus.

```
procedure HPRSE(C)                              ▷ C:Entire Corpus
    sentSet ← extractSent(C)    ▷ sentSet contains entire sentences
 extracted from the corpus
    resultSet ← {}         ▷ sentSet for containing resulting phonetically-rich
 sentences
    while until all the triphones in the corpus are covered do
        triphoneSet ← extractTriphones(sentSet)  ▷ extract triphones
 from all the sentences
        for Each t_i in triphoneSet do
            P_K(t_i) ← Count(t_i) / ∑_{j=1}^{j=m} Count(t_j)
                                         ▷ m : #triphones in the corpus
        end for
        for Each sentence s do
            S(s) ← ∑_{t∈s} 1/P_K(t)              ▷ Score each sentence
            w_1 ← #unqTriphones / #totalTriphone      ▷ for sentence s
            if #triphones in s are too large or too small then
                w_2 ← 0.5
            else
                w_2 ← 1.0
            end if
            S(s) ← S(s) * w_1 * w_2          ▷ Update the sentence score
        end for
        resultSet ← resultSet ∪ bestSent(S)
        sentSet ← sentSet − bestSent(S)
    end while
    return  resultSet ∪ HPBSE(sentSet, resultSet, sentSet ∪
 resultSet, )              ▷ Add the sentences selected from stage 2
end procedure
```

Figure 2. HPRSE's algorithm (Stage-1)

### C. Algorithm

For the purpose of building an efficient speech recognition, the training corpus must have smallest number of grammatically valid as well as phonetically-rich sentences such that the sentence set not only includes all the indispensable recognition

```
procedure HPBSE(S, sentStage1, C)  ▷ S:Remaining sentences in the corpus,
sentStage1:Sentences selected in stage one
    sentSet ← S
    resultSet ← {}
    triphoneStage1 ← extractTriphones(sentStage1)
    for Each t_i in triphoneStage1 do
        P_N(t_i) ← Count(t_i) / Σ_{j=1}^{j=m} Count(t_j)
    end for
    while until the desired number of sentences extracted or sentSet got empty do
        triphoneSet ← extractTriphones(sentSet)
        P_K(t_i) ← Count(t_i) / Σ_{j=1}^{j=m} Count(t_j)
        for Each sentence s in S do
            S(s) ← Σ_{t∈s}[const − const ∗ P_N(t)/P_K(t)]   ▷ Score each sentence
            w_1 ← #unqTriphones / #totalTriphone             ▷ for sentence s
            if #triphones in s are too large or too small then
                w_2 ← 0.5
            else
                w_2 ← 1.0
            end if
            S(s) ← S(s) ∗ w_1 ∗ w_2                           ▷ Update the sentence score
        end for
        if Sim(sentStage1 ∪ bestSent(S(s)), C) ▷ Sim(sentStage1, C)
        then
            resultSet ← resultSet ∪ bestSent(S)
        end if
        sentSet ← sentSet − bestSent(S)
    end while
    return resultSet
end procedure
```

Figure 3. HPBSE's algorithm (Stage-2)

units (triphones), but also these units should be distributed in the same ratio as in the corpus.

A two stage algorithm based on greedy approach is implemented to sort out the phonetically rich sentences from the corpus. Greedy Algorithm is one of the popular way of solving the computer science problems, when a unique optimum solution can not be guaranteed [18]. The greedy approach tries to find out the solution by selecting the locally optimal choices in order to get the global optimum.

The first stage HPRSE (Hindi Phonetically Rich Sentence Extraction) tries to extract minimum set of sentences which cover all possible triphones in the corpus $C$ inputted. The sentences extracted by the stage one are then combined with the sentences extracted by second stage HPBSE (Hindi Phonetically Balanced Sentence Extraction) whose task is to establish the similarity of triphones distribution with the corpus. The approach is being implemented in the form of a two stage algorithm with following conditions taken into consideration:

1) All the triphones present in the corpus must be included in the process of finding phonetically rich sentences (HPRSE algo).
2) The sentences with larger number of distinguished triphones, have to be assigned higher priority.
3) For getting less number of phonetically rich sentences which cover the entire triphones existing in the corpus, the sentences contains triphones of lower frequency in the corpus, should be selected first (HPRSE algo).
4) To match the phonetic distribution in the corpus, the sentences with high frequency of occurrences of triphones have to be selected with higher priority (HPBSE algo).

V. EXPERIMENTS & EVALUATIONS

A. Experimental Details

An experiment is carried out to determine the performance of the proposed approach which is designed to extract a set of phonetically-rich Hindi sentences from the corpus, to be used for developing an automatic Hindi speech recognizer. The experiment is started with converting all the Hindi sentences in the corpus from grapheme representation to phoneme representation by a simple Grapheme2Phoneme (G2P) script written in python. G2P is just a rule-based approach which not only process each word char-by-char but also convert each of them to its phoneme representation simply by mapping according to pair given in table I & II.

Table VI
RESULT OF EXTRACTING PHONETICALLY-RICH SENTENCES FROM EMILLE HINDI CORPUS

| Stage | Selected Sentences | | Selected Triphones | | S=cos($\theta$) |
|---|---|---|---|---|---|
| | Total | % in Corpus | Total | % in Corpus | |
| 1 | 1856 | 0.104 | 34998 | 0.082 | 0.861 |
| 2 | 2500 | 0.140 | 38841 | 0.091 | 0.905 |
| | 3000 | 0.168 | 45243 | 0.106 | 0.932 |
| | 3500 | 0.196 | 50365 | 0.118 | 0.960 |
| | 4000 | 0.224 | 55914 | 0.131 | 0.973 |
| | 4500 | 0.252 | 62316 | 0.146 | 0.977 |
| | 5000 | 0.280 | 68718 | 0.161 | 0.989 |
| | 5500 | 0.308 | 72560 | 0.170 | 0.992 |

The EMILLE Hindi text corpus used here for the experiment contains total of 1,784,784 sentences (42,682,598 triphones). The results of the experiment are summarized in table VI. On observing the results provided in the table VI, It is found that only 1856 sentences extracted in first stage, were able to sufficiently include all the triphones in the corpus. Evidently, the result is signifying the fact that building a phonetically rich corpus doesn't require a very large number of sentences. Here, only 1856 sentences (34,998 triphones) (0.104% of entire sentences and 0.082% of all triphones in the corpus) are proved to be enough to cover all the recognition units.

As for as the distribution of triphones in the extracted sentences is concerned, it is evidently visible that only after the first stage, the distribution (in 1856 sentences) is quite similar to that of the corpus with cosine similarity of 0.861. It is due to the nature of algorithm used in the first stage which selects the sentences consist of triphones of less frequency with high priority.

On the application of second stage, the similarity measure between overall extracted sentences and the entire corpus improves very quickly due to inclusion of more sentences consisting of high frequency triphones. When the total 5500 sentences extracted after finishing the second stage, the statistical distribution is very close to that of the corpus with similarity measure of 0.992 and the percentage of sentences being selected still very low, only 0.308% of the whole corpus.

B. Evaluation

Random selection of sentences is considered here as a baseline for evaluating the performance of the proposed method. Table VII's statistics presents distribution of triphones over the selected sentences through the proposed approach. Correspondingly, table VIII elucidate the case when sentences selected in random order from the corpus. It can be observed from the ratio of unique triphones and total triphones in the table VII, VIII that the proposed method extract the sentences in more uniform way and cover more variety of triphones. Our method is able to extract 8374 unique triphones (42% of all type of triphones in the corpus) on the extraction of 400 sentences. On the other hand a random instance cover only 6380 (32%) unique triphones for the same set of sentences.

Table VII
TRIPHONE EXTRACTED BY PROPOSED METHOD

| #Sentences | #UniqueTriphones | #TotalTriphone | #UnqTriphone/#TotalTriphone |
|---|---|---|---|
| 50 | 552 | 646 | 0.854 |
| 100 | 1058 | 1410 | 0.750 |
| 150 | 2056 | 3264 | 0.629 |
| 200 | 4142 | 6792 | 0.609 |
| 250 | 5165 | 8962 | 0.576 |
| 300 | 6122 | 10956 | 0.558 |
| 350 | 7296 | 13963 | 0.522 |
| 400 | 8374 | 16419 | 0.510 |

Table VIII
TRIPHONE EXTRACTED FROM RANDOMLY SELECTED SENTENCES

| #Sentences | #UniqueTriphones | #TotalTriphone | #UnqTriphone/#TotalTriphone |
|---|---|---|---|
| 50 | 287 | 426 | 0.673 |
| 100 | 725 | 1164 | 0.622 |
| 150 | 1439 | 2698 | 0.533 |
| 200 | 2263 | 4844 | 0.467 |
| 250 | 3144 | 7276 | 0.432 |
| 300 | 4824 | 12529 | 0.385 |
| 350 | 5679 | 16869 | 0.336 |
| 400 | 6380 | 20628 | 0.310 |

## VI. CONCLUSION

It's been a challenging task to decide what should be the input to a speech recognizer to be trained effectively for better and efficient performance. Linguistic experts have suggested that phonetically rich sentences ought to be performed well in training a speech recognizer. The reason is phonetically rich sentences are the minimal set of sentences sufficient enough to train an ASR with optimum set of balanced parameters. Therefore, a novel method has been proposed in this paper to process a corpus in order to extract phonetically-rich sentences based on the triphone distribution for Hindi language. The Hindi corpus used here for the experiment is EMILLE corpus collected & provided by ELRA (European Language Resources Association). All the sentences in the corpus filtered, segmented and converted in the triphone sequences. Afterwords, a two stage algorithm is utilized to extract desired number of phonetically rich and balanced sentences which would be used in training and evaluating an Hindi ASR. The job of first stage is to make sure of uniform distribution of triphones in the selected sentences. The second stage has to match the distribution of triphones in the selected sentences to be balanced with the corpus. The algorithm follow a greedy approach for selecting a sentence with highest score in both the stages. Only difference is that in the second stage the sentence will be included only when it improves triphone distribution on the basis of cosine similarity value between with and without including it. In future, we wish to apply the approach to other Indian languages like Punjabi, Telugu, Marathi etc. We also intended to build an Hindi ASR with the help of already existing language-independent ASR tools e.g. Sphinx[3] and HTK[4] trained on our phonetically-rich sentences.

---

[3] http://cmusphinx.sourceforge.net/
[4] http://htk.eng.cam.ac.uk/